\documentclass{article}

\usepackage{arxiv}

\usepackage[utf8]{inputenc} 
\usepackage[T1]{fontenc}    
\usepackage{hyperref}       
\usepackage{url}            
\usepackage{booktabs}       
\usepackage{amsfonts}       
\usepackage{nicefrac}       
\usepackage{microtype}      
\usepackage{lipsum}		
\usepackage{graphicx}
\usepackage{natbib}
\usepackage{doi}
\usepackage{adjustbox}
\usepackage{multirow}
\usepackage{caption}
\usepackage{float}

\title{Zero-shot generation of synthetic neurosurgical data with large language models}

\author{
    Austin A. Barr\thanks{Correspondence: Austin A. Barr, austin.barr@ucalgary.ca} \\
	Cumming School of Medicine\\
	University of Calgary\\
	Calgary, AB, Canada\\
	\And
	Eddie Guo \\
	Cumming School of Medicine\\
	University of Calgary\\
	Calgary, AB, Canada\\
	\AND
	Emre Sezgin\\
	The Abigail Wexner Research Institute\\
	Nationwide Children’s Hospital\\
	Department of Pediatrics\\
	The Ohio State University College of Medicine\\
	Columbus, OH, United States\\
}

\date{}

\renewcommand{\undertitle}{}
\renewcommand{\shorttitle}{Synthetic Neurosurgical Data}

\hypersetup{
pdftitle={Neurosurgical Synthetic Data},
pdfsubject={cs.CL},
pdfauthor={Austin A. Barr, Eddie Guo, Emre Sezgin},
pdfkeywords={Neurosurgery, Synthetic Data, Large Language Models, Generative Adversarial Networks, Machine Learning}
}

\begin{document}
\maketitle

\begin{abstract}
	\textbf{Objective}: Use of neurosurgical data for clinical research and machine learning (ML) model development is often limited by data availability, sample size, and regulatory constraints. Synthetic data offers a potential solution to challenges associated with accessing and using real-world data (RWD). This study aims to evaluate the capability of zero-shot generation of synthetic neurosurgical data with a large language model (LLM), GPT-4o, by benchmarking data fidelity, utility, and privacy with the conditional tabular generative adversarial network (CTGAN).  \\
    \\
    \textbf{Methods}: A prompt instructed GPT-4o to generate synthetic data based on a plain-language description of the univariate and bivariate statistical properties of a real-world open-access neurosurgical dataset, which included 139 older adults who underwent neurosurgical intervention. The LLM was prompted to generate 10 synthetic datasets matching the original sample size, followed by an additional dataset representing a ten-fold amplification (\textit{n} = 1390). Two datasets were also generated with CTGAN, with sample sizes of 139 and 1390, for benchmarking. Synthetic datasets were compared to the RWD to assess fidelity (means, proportions, distributions, and bivariate correlations), utility (ML classifier performance on RWD), and privacy (duplication of records from RWD). \\
    \\
    \textbf{Results}: The GPT-4o-generated datasets matched or exceeded CTGAN performance, despite no fine-tuning or access to RWD for pre-training. Datasets demonstrated high univariate and bivariate fidelity to RWD without directly exposing any real patient records. Training an ML classifier on the GPT-4o-generated dataset (\textit{n} = 1390) and testing on RWD for a binary prediction task showed an F1 score (0.706) with comparable performance to training on the CTGAN data (0.705) for predicting postoperative functional status deterioration.\\
    \\
    \textbf{Conclusions}: GPT-4o demonstrated a promising ability to generate high-fidelity synthetic neurosurgical data. These findings also indicate that data synthesized with GPT-4o can effectively augment clinical data with small sample sizes, and train ML models for prediction of neurosurgical outcomes. Further investigation is necessary to improve the preservation of distributional characteristics and boost classifier performance.
\end{abstract}

\keywords{Neurosurgery \and Synthetic Data \and Large Language Models \and Generative Adversarial Networks \and Machine Learning}

\section{Introduction}
Neurosurgical research relies on data collected from research initiatives or routinely during surgical and perioperative care to identify disease patterns, evaluate interventions, and assess patient outcomes. High-quality neurosurgical data is also increasingly applied toward the training and validation of machine learning (ML) models for outcome prediction, image analysis, and operative planning [1-3]. However, there are significant barriers to acquiring, sharing, and using real-world data (RWD). Use of clinical data is constrained by data availability, incomplete data, small sample sizes, privacy regulations, and resource-intensive preprocessing and de-identification procedures [4-6]. The relative rarity of many intracranial diseases and limited patient volumes at individual institutions [2,7] pose barriers to assembling comprehensive and high-quality datasets for meaningful analyses. Routinely captured data is often missing information [5], either during care or when patients are lost to follow-up, which may limit the evaluation of long-term outcomes in procedural care. Requirements for accessing (e.g., institutional review boards, data privacy regulations), sharing (e.g., data-sharing agreements), and using (e.g., pre-processing, de-identification) clinical data are necessary to maintain patient privacy, yet costly, resource-intensive, and often inaccessible for researchers [8,9]. Processes to obtain data-sharing agreements restrict cross-institutional research collaboration [10], which is often necessary to assemble large neurosurgical datasets. Beyond the time and resources necessary for data preprocessing, de-identification procedures often remove or obscure information which may be relevant for analyses (e.g., demographics, geographics, temporal factors).\\
\\
To address these challenges, several initiatives have been undertaken to facilitate data sharing and collaboration among neurosurgical researchers. The American Spine Registry, Stereotactic Radiosurgery Registry, and the Quality Outcomes Database (QOD) including the NeuroVascular Quality Initiative-Quality Outcomes Database (NVQI-QOD) and the QOD Tumor Registry [11-13], host clinical data from participating institutions across the United States. While these registries have made significant contributions [14,15], they still face limitations in international collaboration and scope. Clinical registries focus on specific subdisciplines, which may not include all areas of neurosurgery and perioperative care. Collectively, these factors emphasize the importance of exploring alternative approaches to provide researchers with high-quality neurosurgical data while maintaining patient privacy.\\
\\
Synthetic data has emerged as a promising solution to challenges associated with data scarcity and privacy. This approach involves the creation of artificial datasets which retain statistical properties and relationships within RWD, but are designed to preserve patient privacy and enable greater opportunities for data sharing and use [16,17]. Current methods of creating synthetic data predominantly involve generative adversarial networks (GANs) and variational autoencoders (VAEs) [18,19]. While GANs and VAEs have demonstrated significant fidelity and utility of synthetic imaging and tabular data [20-22], several limitations persist. These approaches require technical expertise, access to RWD for training, and are associated with several privacy concerns—particularly when trained on smaller datasets [23]. Recent results have emerged in generating synthetic tabular data using another form of generative artificial intelligence: large language models (LLMs). Using zero-shot plain-language prompting, the LLM GPT-4o was capable of generating synthetic data which retained univariate statistical properties and simple between-parameter relationships found in real-world perioperative data [24]. While promising, questions remained regarding the utility of data for training ML models, capacity for data enhancement (i.e., amplification, augmentation), and applicability toward neurosurgical data.\\
\\
The present study aims to expand upon existing research which has evaluated a zero-shot approach to synthetic tabular data generation using GPT-4o. Specifically, we evaluate the fidelity, utility, and privacy of synthetic neurosurgical data generated with GPT-4o, and benchmark performance against a GAN designed for tabular data synthesis: conditional tabular generative adversarial network (CTGAN) [25]. Comparisons in means, proportions, distributions, bivariate correlations, ML model performance, and uniqueness of records are made between synthetic and RWD. These metrics are also evaluated with amplified sample sizes and with the addition of newly synthesized, interrelated features. By focusing on a small neurosurgical dataset that encompasses a variety of pathologies and perioperative parameters [26], this work seeks to demonstrate the applicability of LLMs toward addressing challenges in data access and availability for neurosurgical research.

\section{Methods}

\subsection{Real-World Dataset and Parameter Selection}
The real-world neurosurgical dataset used as a comparator in this study was prospectively collected from a single high-volume tertiary care center in Milan, Italy [26,27]. It included records of 143 patients aged 65 years or older, undergoing various neurosurgical interventions between 1st January, 2018 and 31st December, 2019. This dataset was chosen for its breadth of clinically relevant parameters.\\
\\
The selection of parameters for the present analysis was guided by their clinical significance, completeness, and to reflect a diversity of data formats (i.e., numerical, text), variable types (i.e., continuous, categorical, ordinal, binary), distributions (i.e., normal, skewed), and measures relevant to perioperative assessment. A total of 12 parameters were selected, and are outlined in Table \ref{tab:neurosurgical_dataset_summary}.

\begin{table}[H]
    \captionsetup{width=0.8\linewidth}
    \caption{Summary of selected parameters from the real-world neurosurgical dataset.\\}
	\centering
    \centering
    \begin{tabular}{|l|l|}
    \hline
    \textbf{Category (\textit{n})} & \textbf{Parameters (units)} \\ \hline
    Demographic data (2) & Age (years), biological sex (M/F) \\ \hline
    Preoperative measures (5) & 
    \begin{tabular}[c]{@{}l@{}} 
    Body mass index (kg/m\textsuperscript{2}), American Society of \\ 
    Anesthesiologists (ASA) physical status classification (1-6), \\ 
    heart disease (yes/no), diabetes (yes/no), previous \\ 
    brain radiotherapy (yes/no) 
    \end{tabular} \\ \hline
    Surgical parameters (2) & Histology, Milan complexity scale (MCS) (0-8) \\ \hline
    Postoperative outcomes (3) & 
    \begin{tabular}[c]{@{}l@{}} 
    Postoperative length of stay (days), Karnofsky \\ 
    performance status (KPS) deterioration at discharge \\ 
    (yes/no), Landriel scale (0-4) 
    \end{tabular} \\ \hline
    \end{tabular}
    \label{tab:neurosurgical_dataset_summary}
\end{table}

\subsection{Data Cleaning and Preparation}
From the initial 143 patients, a total of 139 case files were included. Patients were removed for erroneous length of stay (LOS) duration (\textit{n} = 3) and missing histology classification (\textit{n} = 1). Histological classifications were divided into 20 distinct categories, postoperative LOS was calculated as the duration from date of surgery to date of discharge, Karnofsky performance status (KPS) deterioration was dichotomized by calculating the difference between KPS preoperatively and at discharge, Milan complexity scale (MCS) was calculated from binary intraoperative variables (i.e., major brain vessel manipulation, posterior fossa, cranial nerve manipulation, eloquent area, tumor size), and the Landriel scale was categorized into scores between 0 and 4.\\
\\
Following parameter selection and data preprocessing, univariate statistics were calculated to summarize continuous parameters (mean, standard deviation, and range) and categorical/ordinal/binary parameters (proportions). Histograms of continuous parameters were inspected for normality. All bivariate correlations between continuous, ordinal, and binary parameters were calculated using Pearson’s product-moment correlation. Any correlation demonstrating a moderate or stronger relationship of |\textit{r}| > 0.5 was noted. The relationship between KPS deterioration at discharge and Landriel scale (\textit{r} = 0.57) was the only bivariate relationship to meet this threshold, consistent with the original study’s primary outcome [27].

\subsection{Generation of Synthetic Datasets Using GPT-4o}
\begin{figure}[H]
    \centering
    \includegraphics[width=0.75\textwidth]{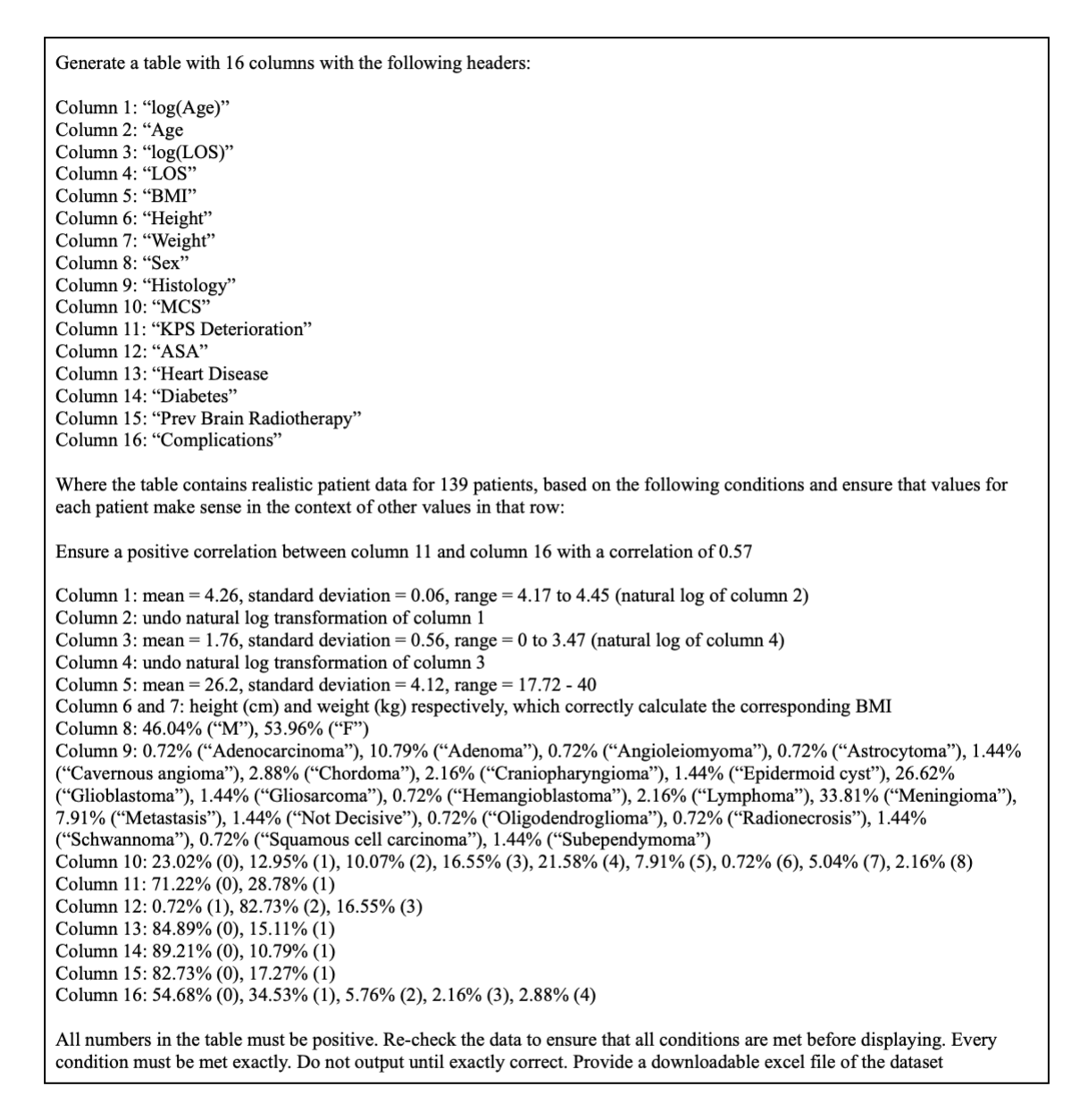}
    \captionsetup{width=0.8\linewidth}
    \caption{Prompt inputted into GPT-4o to generate synthetic datasets. The prompt was inputted over 10 independent trials to generate synthetic datasets (\textit{n} = 139). To generate an amplified dataset, the prompt was modified from “139 patients” to “1390 patients”, with all other aspects of the prompt remaining identical.}
    \label{fig:GPT_prompt}
\end{figure}

To generate synthetic data with GPT-4o, a zero-shot prompting approach was used. Univariate and bivariate statistical properties of RWD were inputted into GPT-4o along with instructions on desired structure of the tabular dataset (i.e., organization, column names), in a plain-language prompt, provided in Figure \ref{fig:GPT_prompt}. For continuous parameters, the prompt included mean, standard deviation, and range. Natural log transformations were applied to the patient age and postoperative LOS parameters to normalize skewed distributions, and the prompt included properties of the transformed values. The prompt also included instructions to generate separate columns for age and postoperative LOS with removed log transformations. Based on each case file’s body mass index (BMI), the model was instructed to generate new and corresponding height and weight features. Height and weight measurements were not included in RWD and no formulas were provided to guide these calculations. For categorical, ordinal, and binary parameters, proportions were provided. The prompt also included the correlation coefficient between KPS deterioration at discharge and Landriel scale (\textit{r} = 0.57), and was instructed to maintain this bivariate relationship. Collectively, a total of 16 columns were necessary to output.\\

Prior to data generation, no pre-training or task-specific fine-tuning was performed, and no patient data was provided in the prompt. The zero-shot prompting approach involved instructing GPT-4o to generate a dataset of the same size as the reference dataset (\textit{n} = 139), outputted in a downloadable spreadsheet format (.xlsx). This prompt was inputted into GPT-4o over 10 independent trials (new session with memory disabled) to assess the reproducibility of results. Following, GPT-4o was instructed to generate an additional dataset (\textit{n} = 1,390), representing a ten-fold increase in sample size. The only adjustment made to the prompt was replacing “139 patients” with “1390 patients.”

\subsection{Generation of Synthetic Datasets Using CTGAN}
To generate synthetic data with CTGAN, data was preprocessed to handle different variable types. Continuous parameters were normalized using the MinMaxScaler, from the Python library skikit-learn\footnote{\href{https://scikit-learn.org/}{https://scikit-learn.org/}}, which transformed the data to a range between 0 and 1. Ordinal and binary parameters were converted to integer values to ensure an inherent order of categories, and the categorical histology parameter was encoded, using the LabelEncoder, to convert a string of labels into integers, allowing CTGAN to handle them as discrete values. \\
\\
To determine the optimal number of training epochs, loss functions were analyzed. The loss function plateaued after 2500 epochs, with only minor fluctuations thereafter, indicating no significant improvement in model performance. Based on this observation, 3000 epochs was chosen to avoid overfitting and ensure model stability. Following, a dataset with the same number of samples as the RWD (\textit{n} = 139) and an amplified dataset (\textit{n} = 1390) were generated. Once generated, transformations applied during preprocessing were removed.

\subsection{Statistical Analysis}
Each generated dataset was assessed for output errors and calculations performed by GPT-4o were validated. Evaluation of output errors ensured each dataset was structured in the desired tabular format, columns were appropriately labelled, and the dataset was complete. Calculations performed by GPT-4o were validated, including the relationship between generated height, weight, and BMI parameters for each case file as well as the removal of log transformations for the age and postoperative LOS parameters. These calculations were assessed following rounding to the nearest hundredth.\\
\\
The fidelity, utility, and privacy of synthetic datasets were quantified using the open-source Python library Synthetic Data Metrics (SDMetrics)\footnote{\href{https://docs.sdv.dev/sdmetrics}{https://docs.sdv.dev/sdmetrics}}, which measures performance on a normalized scale from 0 to 1. The metrics used in the present analysis are summarized in Table \ref{tab:synthetic_data_metrics}. Benchmarking was performed between GPT-4o and CTGAN-generated datasets of equivalent sample size. The StatisticSimilarity and KSComplement metrics were performed on continuous parameters, TVComplement was performed on categorical, ordinal, and binary parameters, CorrelationSimilarity was performed on the relationship between KPS deterioration at discharge and Landriel scale for each dataset, Binary Classification was performed on the amplified datasets (\textit{n} = 1390), and NewRowSynthesis was performed on all datasets.

\begin{table}[H]
    \centering
    \captionsetup{width=0.8\linewidth}
    \caption{Metrics used to evaluate the fidelity, utility, and privacy of synthetic data. All metrics are included in the Synthetic Data Metrics (SDMetrics) open-source Python library. Names of metrics/functions are listed as they appear in the library.}

    \begin{tabular}{|p{3cm}|p{3cm}|p{7cm}|}
        \hline
        \textbf{Metrics/Functions} & \textbf{Type of Assessment} & \textbf{Summary} \\ \hline
        StatisticSimilarity       & \centering Fidelity                    & Similarity in means between real and synthetic datasets \\ \hline
        \vfill KSComplement              & \vfill \centering Fidelity                    & Distributional similarity between numerical data in real and synthetic datasets (Kolmogorov-Smirnov test) \\ \hline
        \vfill TVComplement              & \vfill \centering Fidelity                    & Distributional similarity between categorical, ordinal, and binary data in real and synthetic datasets (total variation distance) \\ \hline
        \vfill CorrelationSimilarity     & \vfill \centering Fidelity                    & Similarity of bivariate correlations in real and synthetic datasets (Pearson’s product moment correlation) \\ \hline
        \vfill Binary Classification     & \vfill \centering Utility                     & Performance of training a machine learning classifier (BinaryAdaBoostClassifier) on synthetic data to perform a binary prediction task on a real dataset (train-synthetic-test real (TSTR)) \\ \hline
        \vfill NewRowSynthesis           & \vfill \centering Privacy                     & Uniqueness of rows in synthetic data compared to a real dataset \\ \hline
    \end{tabular}
    \label{tab:synthetic_data_metrics}
\end{table}

A train-synthetic-test-real (TSTR) framework was used to assess the utility of the amplified synthetic datasets (\textit{n} = 1390). In two independent evaluations, synthetic data was used to train an ML classifier (BinaryAdaBoostClassifier) to perform a binary classification task on the RWD. The histology parameter was excluded due to the model’s difficulty with non-numeric categorical data. The classifier was trained on synthetic data and the binarized KPS deterioration at discharge parameter served as the target. After training, the model’s class probabilities for the positive class (class 1) were calculated on the RWD. These probabilities were used to test various classification thresholds, ranging from 0 to 1 (in increments of 0.01). For each threshold, predictions were made by assigning class 1 if the probability exceeded the threshold, and class 0 otherwise. Performance at each threshold was evaluated by calculating the confusion matrix and key classification metrics (precision, recall, F1 score). The threshold that provided the best performance, along with the corresponding metrics, were recorded and compared to the performance of the classifier trained on the other synthetic dataset.\\
\\
For further evaluation of fidelity, comparisons in 95\% confidence intervals (CI) for continuous parameters were visualized with error bars. Proportional alignment of ordinal and binary data was visualized using stacked bar plots, and a summary table for histology. For the ten GPT-4o-generated datasets with the same sample size as the real dataset, performance was ranked to display the greatest and poorest performing generation for each parameter. Continuous parameters were ordered by 95\% CI overlap percentage with the RWD. Categorical, ordinal, and binary parameters were ordered by TVComplement score. Best and worst performing generations for the given parameter were displayed. For any ties in TVComplement performance with distinct category proportions between generations, the earlier generated dataset was displayed.\\
\\
Statistical testing, beyond SDMetrics, was performed using R statistical software (version 4)\footnote{\href{https://www.r-project.org}{https://www.r-project.org}}. Figures were generated using the Python library Matplotlib\footnote{\href{https://matplotlib.org}{https://matplotlib.org}}.

\subsection{Ethical Considerations}
The real-world neurosurgical dataset was used in accordance with Creative Commons Attribution 4.0 International. No identifiable patient data was used in this study, no real patient data was inputted into GPT-4o for pre-training or during dataset generation, and no real patient data has been included in this paper. The synthetic datasets were generated and evaluated solely for the present study.

\section{Results}
\subsection{Validation and New Feature Synthesis}
Over successive trials, GPT-4o generated 11 complete and structured tabular datasets that aligned with the prompted instructions. Validation of calculations performed by GPT-4o demonstrated no calculation errors in BMI and removal of natural log transformations. The BMI corresponded appropriately to the generated height and weight parameters for each case file in every generated dataset. The generated heights and weights for all generations were realistic, with means of heights ranging from 166.74 to 174.87 centimeters and weights from 74.46 to 80.61 kilograms. The age and postoperative LOS parameters were correctly calculated from log-transformed values for all case files, across the 11 generated datasets.
\subsection{Synthetic Data Fidelity and Privacy}
\begin{figure}[H]
    \centering
    \includegraphics[width=0.75\textwidth]{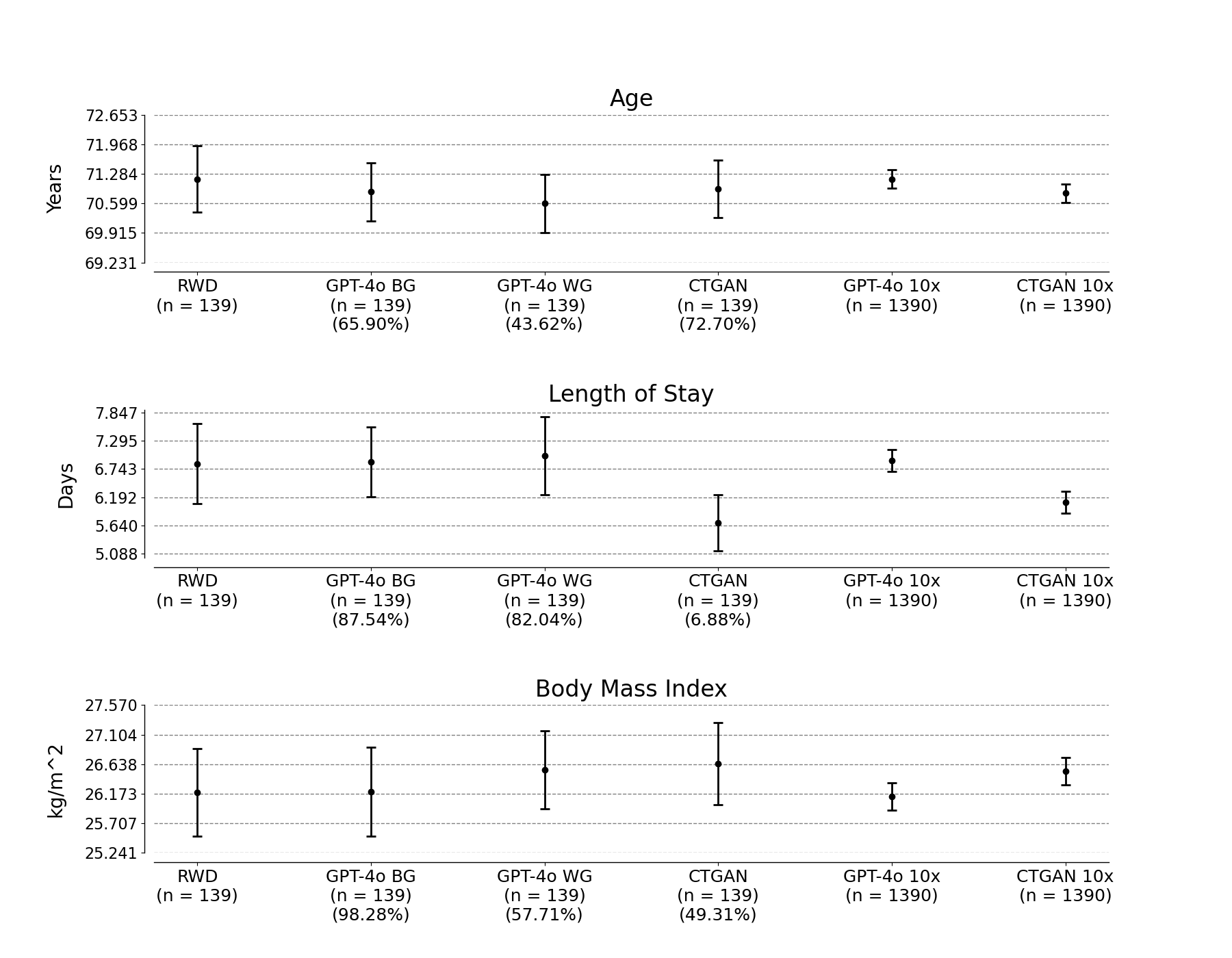}
    \captionsetup{width=0.8\linewidth}
    \caption{95\% confidence interval (CI) overlap of continuous parameters between real and synthetic datasets. Error bars display 95\% CIs for age, postoperative length of stay, and body mass index for the real-world data (RWD), best (BG) and worst (WG) generations from GPT-4o (\textit{n} = 139) datasets, CTGAN (\textit{n} = 139) dataset, and ten-fold amplified datasets (\textit{n} = 1390) generated with GPT-4o and CTGAN (10x). The GPT-4o (\textit{n} = 139) datasets were ranked in order of 95\% CI overlap percentage to determine BG and WG. The 95\% CI overlap percentages were not included for 10x datasets due to smaller 95\% CIs with larger sample size.}
    \label{fig:CI_overlap}
\end{figure}

\begin{figure}[H]
    \centering
    \includegraphics[width=0.75\textwidth]{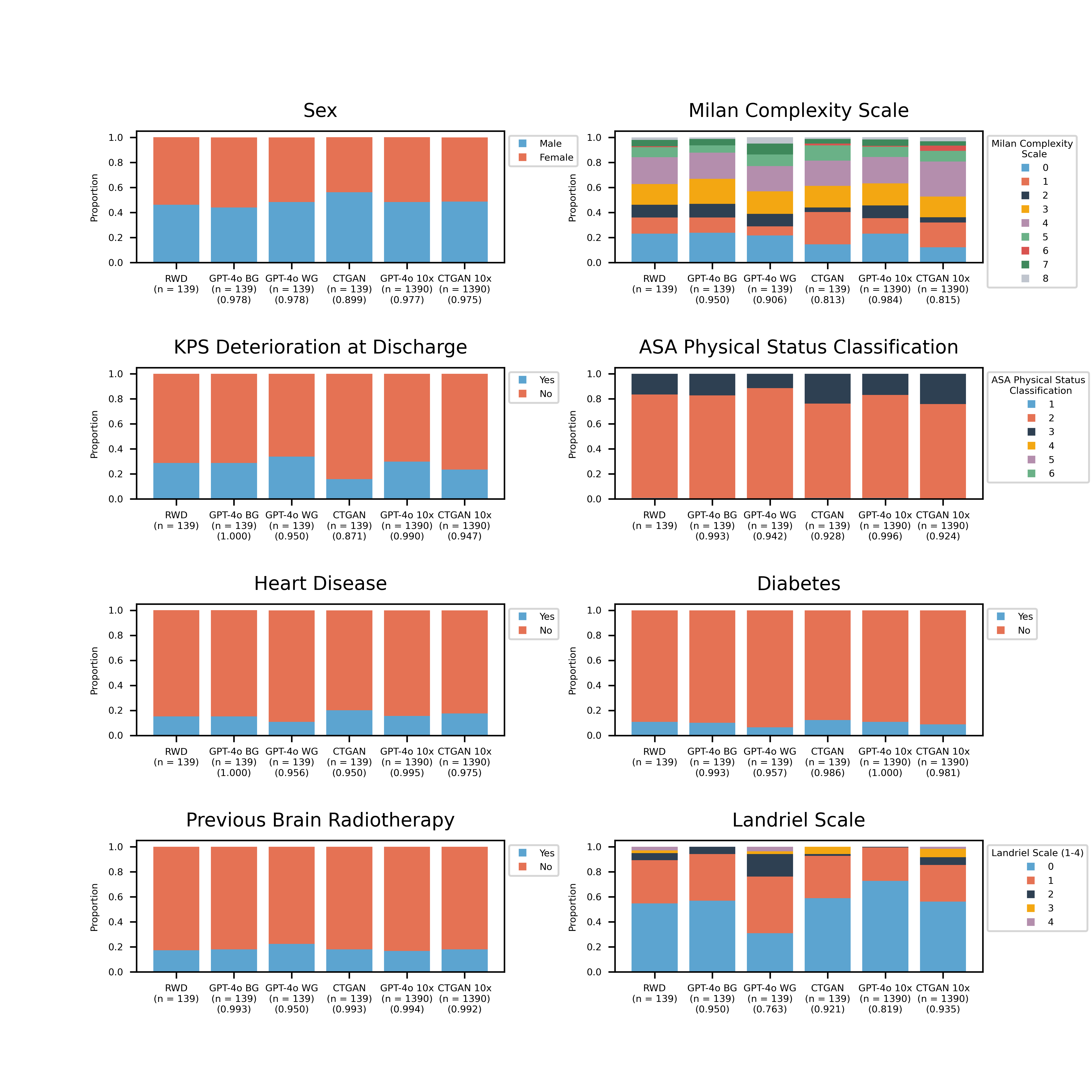}
    \captionsetup{width=0.8\linewidth}
    \caption{Proportional alignment of ordinal and binary parameters between real and synthetic datasets. Stacked bar plots display proportions of sex, Karnofsky performance status (KPS) deterioration at discharge, heart disease, previous brain radiotherapy, Milan complexity scale, American Society of Anesthesiologists (ASA) physical status classification, diabetes, and Landriel scale for the real-world data (RWD), best (BG) and worst (WG) generations from GPT-4o (\textit{n} = 139) datasets, CTGAN (\textit{n} = 139) dataset, and ten-fold amplified datasets (\textit{n} = 1390) generated with GPT-4o and CTGAN (10x). The GPT-4o (\textit{n} = 139) datasets were ranked in order of TVComplement score to determine BG and WG.}
    \label{fig:proportions}
\end{figure}

Preservation of means from GPT-4o-generated data consistently outperformed CTGAN-generated data across all generations, as seen in Table \ref{tab:sdmetrics_results}. Moderately higher 95\% CI overlap for (\textit{n} = 139) datasets was observed in the CTGAN-generated age parameter, and greater overlap was observed in the postoperative LOS and BMI parameters generated with GPT-4o, displayed in Figure \ref{fig:CI_overlap}. Overall preservation of continuous parameters’ distributional characteristics was limited. The GPT-4o-generated (\textit{n} = 139) datasets demonstrated greater overall fidelity compared to the benchmarked CTGAN dataset, and poorer performance on amplified data compared to CTGAN. Distributional characteristics of the categorical, ordinal, and binary parameters were better preserved by GPT-4o across all generations, displayed in Figure \ref{fig:proportions} and Tables \ref{tab:sdmetrics_results} and \ref{tab:histology}.

\begin{table}[H]
    \centering
    \small
    \captionsetup{width=0.8\linewidth}
    \caption{Evaluation of synthetic data fidelity, utility, and privacy. Comparisons in performance are made between synthetic data generations with equivalent sample sizes, with bolded values as the best performing generation for the respective metric. BG, best generation; WG, worst generation; 10x, ten-fold amplification.}
    \begin{adjustbox}{max width=\textwidth}
    \begin{tabular}{|l|c|c|c|c||c|c|}
        \hline
        \multirow{2}{*}{\textbf{Metric}} 
        & \multicolumn{1}{|c|}{\textbf{GPT-4o BG}} 
        & \multicolumn{1}{|c|}{\textbf{GPT-4o WG}} 
        & \multicolumn{1}{|c|}{\textbf{GPT-4o Average}} 
        & \multicolumn{1}{|c||}{\textbf{CTGAN}} 
        & \multicolumn{1}{|c|}{\textbf{GPT-4o 10x}} 
        & \multicolumn{1}{|c|}{\textbf{CTGAN 10x}} \\
        & ($n=139$) & ($n=139$) & ($n=139$) & ($n=139$) & ($n=1390$) & ($n=1390$) \\ 
        \hline
        \multicolumn{7}{|l|}{\textbf{Fidelity}} \\
        \hline
        StatisticSimilarity & \textbf{0.995$\pm$0.008} & 0.984$\pm$0.011 & 0.987$\pm$0.008 & 0.977$\pm$0.013 & \textbf{0.998$\pm$0.002} & 0.981$\pm$0.006 \\
        KSComplement & \textbf{0.885$\pm$0.022} & 0.863$\pm$0.056 & 0.880$\pm$0.054 & 0.878$\pm$0.062 & 0.889$\pm$0.042 & \textbf{0.934$\pm$0.038} \\
        TVComplement & \textbf{0.965$\pm$0.032} & 0.937$\pm$0.068 & 0.949$\pm$0.051 & 0.886$\pm$0.117 & \textbf{0.965$\pm$0.059} & 0.911$\pm$0.109 \\
        CorrelationSimilarity$^*$ & \textbf{0.996} & 0.805 & 0.920$\pm$0.060 & 0.880 & 0.945 & \textbf{0.990} \\
        \hline
        \multicolumn{7}{|l|}{\textbf{Utility}} \\
        \hline
        TSTR (F1) & - & - & - & - & \textbf{0.706} & 0.705 \\
        \hline
        \multicolumn{7}{|l|}{\textbf{Privacy}} \\
        \hline
        NewRow Synthesis & \textbf{1.000} & \textbf{1.000} & \textbf{1.000$\pm$0} & \textbf{1.000} & \textbf{1.000} & \textbf{1.000} \\
        \hline
    \end{tabular}
    \end{adjustbox}
    
    \vspace{2mm}
    \raggedright{\footnotesize $^*$KPS Deterioration and Landriel Scale}
    \label{tab:sdmetrics_results}
\end{table}

Analysis of the bivariate relationship between KPS deterioration at discharge and the Landriel scale, demonstrated that all generated datasets preserved a positive correlation. Compared to the moderate positive correlation (\textit{r} = 0.57) in RWD, the GPT-4o-generated (\textit{n} = 139) datasets had 6/10 (60.0\%) generations preserve a moderate positive correlation (\textit{r} = 0.40 to 0.58), 2/10 (20.0\%) with a strong positive correlation (\textit{r} = 0.63 to 0.77), 1/10 (10.0\%) with a weak positive correlation (\textit{r} = 0.27), and 1/10 (10.0\%) with a very weak positive correlation (\textit{r} = 0.18). The best-preserved positive correlation was observed in generations 3 and 4, both with a moderate positive correlation (\textit{r} = 0.58). Across these 10 GPT-4o-generations, the mean and range of generated correlations was \textit{r} = 0.47 (0.18 - 0.77). The CTGAN-generated (\textit{n} = 139) dataset displayed a weak positive correlation (\textit{r} = 0.33), which was outperformed by 8/10 (80.0\%) of the GPT-4o-generated datasets. The correlation associated with the amplified dataset generated with CTGAN (\textit{r} = 0.55) demonstrated better preservation of the bivariate relationship compared to the GPT-4o-generated amplified dataset (\textit{r} = 0.46).

All generations did not include any exact row matches to the RWD, which was assessed using the NewRowSynthesis metric.

The generated datasets are available at our GitHub repository\footnote{\href{https://github.com/aabarr/Synthetic-Neurosurgical-Data}{https://github.com/aabarr/Synthetic-Neurosurgical-Data}}.

\begin{table}[H]
    \centering
    \small
    \captionsetup{width=0.8\linewidth}
    \caption{Proportional alignment of the histology parameter between real and synthetic datasets. Comparisons in number of observations of each categorization of histology between the real-world data (RWD), best (BG) and worst (WG) generations from GPT-4o (\textit{n} = 139) datasets, CTGAN (\textit{n} = 139) dataset, and ten-fold amplified datasets (\textit{n} = 1390) generated with GPT-4o and CTGAN (10x). The GPT-4o (\textit{n} = 139) datasets were ranked in order of TVComplement score to determine BG and WG.}
    \begin{adjustbox}{max width=\textwidth}
    \begin{tabular}{|l|c|c|c|c||c|c|}
        \hline
        \multirow{2}{*}{\textbf{Parameter/Metric}} 
        & \multicolumn{1}{|c|}{\textbf{Real}} 
        & \multicolumn{1}{|c|}{\textbf{GPT-4o BG}} 
        & \multicolumn{1}{|c|}{\textbf{GPT-4o WG}} 
        & \multicolumn{1}{|c||}{\textbf{CTGAN}} 
        & \multicolumn{1}{|c|}{\textbf{GPT-4o 10x}} 
        & \multicolumn{1}{|c|}{\textbf{CTGAN 10x}} \\
        & ($n=139$) & ($n=139$) & ($n=139$) & ($n=139$) & ($n=1390$) & ($n=1390$) \\ 
        \hline
        \multicolumn{7}{|l|}{\textbf{Histology}} \\
        \hline
        Adenocarcinoma & 1 (0.72\%) & 0 (0.0\%) & 2 (1.44\%) & 3 (2.16\%) & 10 (0.72\%) & 28 (2.01\%) \\
        Adenoma & 15 (10.79\%) & 18 (12.95\%) & 15 (10.79\%) & 14 (10.07\%) & 140 (10.07\%) & 139 (10.00\%) \\
        Angioleiomyoma & 1 (0.72\%) & 1 (0.72\%) & 2 (1.44\%) & 4 (2.88\%) & 5 (0.36\%) & 35 (2.52\%) \\
        Astrocytoma & 1 (0.72\%) & 0 (0.0\%) & 1 (0.72\%) & 2 (1.44\%) & 16 (1.15\%) & 33 (2.37\%) \\
        Cavernous angioma & 2 (1.44\%) & 4 (2.88\%) & 1 (0.72\%) & 7 (5.04\%) & 23 (1.65\%) & 46 (3.31\%) \\
        Chordoma & 4 (2.88\%) & 5 (3.60\%) & 4 (2.88\%) & 13 (9.35\%) & 37 (2.66\%) & 71 (5.11\%) \\
        Craniopharyngioma & 3 (2.16\%) & 7 (5.04\%) & 4 (2.88\%) & 4 (2.88\%) & 24 (1.73\%) & 63 (4.53\%) \\
        Epidermoid cyst & 2 (1.44\%) & 4 (2.88\%) & 4 (2.88\%) & 12 (8.63\%) & 22 (1.58\%) & 55 (3.96\%) \\
        Glioblastoma & 37 (26.62\%) & 39 (28.06\%) & 30 (21.58\%) & 17 (12.23\%) & 400 (28.78\%) & 175 (12.59\%) \\
        Gliosarcoma & 2 (1.44\%) & 1 (0.72\%) & 0 (0.0\%) & 6 (4.32\%) & 21 (1.51\%) & 49 (3.53\%) \\
        Hemangioblastoma & 1 (0.72\%) & 0 (0.0\%) & 0 (0.0\%) & 0 (0.0\%) & 11 (0.79\%) & 31 (2.23\%) \\
        Lymphoma & 3 (2.16\%) & 2 (1.44\%) & 3 (2.16\%) & 8 (5.76\%) & 41 (2.95\%) & 59 (4.24\%) \\
        Meningioma & 47 (33.81\%) & 42 (30.22\%) & 56 (40.29\%) & 15 (10.79\%) & 460 (33.09\%) & 200 (14.39\%) \\
        Metastasis & 11 (7.91\%) & 11 (7.91\%) & 7 (5.04\%) & 12 (8.63\%) & 99 (7.12\%) & 127 (9.14\%) \\
        Not decisive & 2 (1.44\%) & 2 (1.44\%) & 2 (1.44\%) & 3 (2.16\%) & 23 (1.65\%) & 67 (4.82\%) \\
        Oligodendroglioma & 1 (0.72\%) & 0 (0.0\%) & 1 (0.72\%) & 1 (0.72\%) & 9 (0.65\%) & 39 (2.81\%) \\
        Radionecrosis & 1 (0.72\%) & 1 (0.72\%) & 1 (0.72\%) & 4 (2.88\%) & 14 (1.01\%) & 34 (2.45\%) \\
        Schwannoma & 2 (1.44\%) & 0 (0.0\%) & 2 (1.44\%) & 2 (1.44\%) & 14 (1.01\%) & 59 (4.24\%) \\
        Squamous cell carcinoma & 1 (0.72\%) & 1 (0.72\%) & 2 (1.44\%) & 6 (4.32\%) & 7 (0.50\%) & 36 (2.59\%) \\
        Subependymoma & 2 (1.44\%) & 1 (0.72\%) & 2 (1.44\%) & 6 (4.32\%) & 14 (1.01\%) & 44 (3.17\%) \\
        \hline
        \textbf{TVComplement} & - & \textbf{0.899} & 0.892 & 0.612 & \textbf{0.927} & 0.658 \\
        \hline
    \end{tabular}
    \end{adjustbox}
    \label{tab:histology}
\end{table}

\subsection{Synthetic Data Utility}
\begin{figure}[H]
    \centering
    \includegraphics[width=0.75\textwidth]{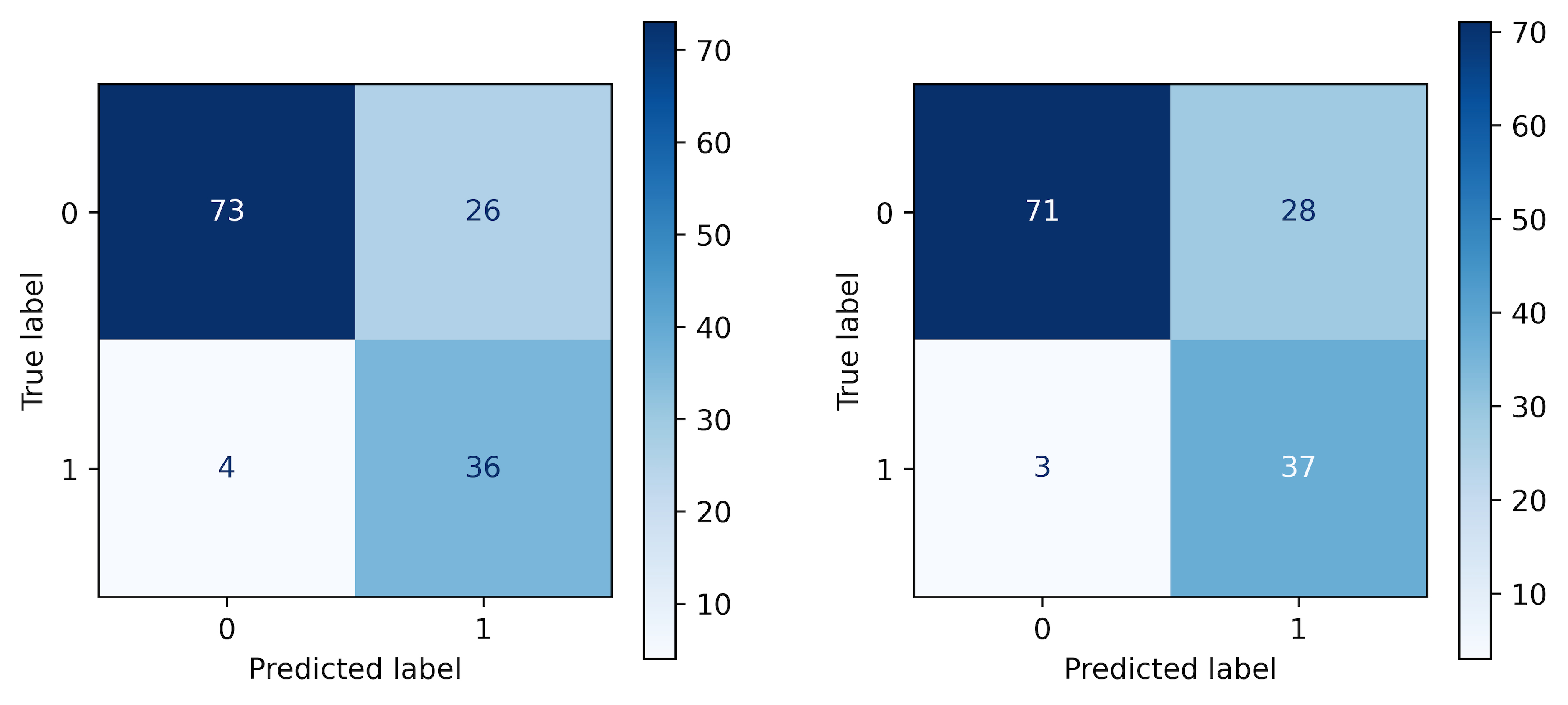}
    \captionsetup{width=0.8\linewidth}
    \caption{Confusion matrices for the binary classification task trained on amplified synthetic data and tested on real data. Binary classification used the BinaryAdaBoostClassifier to predict Karnofsky performance status (KPS) deterioration at discharge. The confusion matrices display performance of the (left) model trained on the GPT-4o (\textit{n} = 1390) dataset and (right) model trained on the CTGAN (\textit{n} = 1390) dataset.}
    \label{fig:confusion_matrices}
\end{figure}

The TSTR binary classification analysis demonstrated comparable data utility between the amplified datasets generated with GPT-4o and CTGAN. The F1 scores for predicting KPS deterioration at discharge on RWD were 0.706 (GPT-4o) and 0.705 (CTGAN), listed in Table \ref{tab:sdmetrics_results}. The threshold yielding the best performance for GPT-4o data was 0.49, with a precision of 0.581, and recall of 0.900. The threshold yielding the best performance for CTGAN data was 0.40, with a precision of 0.569, and recall of 0.925. The confusion matrices associated with training on each dataset and testing on RWD are displayed in Figure \ref{fig:confusion_matrices}.

\newpage

\section{Discussion}
Herein we present a novel approach to generating synthetic tabular neurosurgical data with GPT-4o. Using a zero-shot approach, with a plain-language prompt describing the desired dataset structure and statistical properties, GPT-4o was capable of reproducibly generating data with high fidelity to RWD. This approach was also shown to be effective in ten-fold amplification of a small neurosurgical dataset, generating new and interrelated features, and performing calculations on parameters without provided formulae. Generated datasets included a variety of perioperative parameters, data types, and histology, suggesting broad applicability toward synthetic neurosurgical data generation. \\
\\
Benchmarking GPT-4o demonstrated that an LLM-based approach to synthetic data generation matched or exceeded CTGAN performance across several fidelity, utility, and privacy metrics. Notably, the binary classifier trained on an amplified dataset generated with GPT-4o comparably performed to a model trained on CTGAN-generated data for predicting postoperative functional status deterioration on RWD. This provides promising initial evidence of LLM-generated synthetic data utility. In line with our previous work on zero-shot generation of tabular data, our findings also demonstrate the preservation of between-parameter relationships [24]. However, the present results also show preserved strength of correlation, interrelations between new features (i.e., height, weight), and ten-fold amplification of sample size. Increasing sample size, while retaining statistical properties of RWD, is of particular value to neurosurgical research, where small sample sizes often limit the robustness of statistical analyses and development of ML models [2,7].\\
\\
The zero-shot approach, definitionally, does not require pre-training. Therefore, limitations surrounding data access and privacy, which are associated with other synthetic data generation methods, are not of comparable concern. Furthermore, GANs may memorize individual training data, leaving the model vulnerable to membership inference attacks [23,28]. These models are especially susceptible to attacks when trained on smaller datasets, with research suggesting that this vulnerability presents greater risk if training data falls well below 10,000 samples [23]. This presents an additional challenge in applying GANs/VAEs to neurosurgical data, as it is unlikely that sample sizes will reach sufficient scale to prevent these attacks using current approaches. The presented zero-shot approach also allows researchers to specify the desired dataset structure and statistical properties using a plain-language prompt, which can eliminate the technical expertise and fine-tuning necessary to generate synthetic data using GANs and VAEs. The model outputs a downloadable and ready-to-use spreadsheet, bypassing resource-intensive preprocessing and de-identification procedures required for use of RWD [8]. Further, because data can be amplified and publicly shared, this may overcome current challenges associated with neurosurgical research (i.e., data scarcity, incomplete datasets, small sample sizes, regulatory requirements), and enable greater possibility for data access, cross-institutional collaboration, and scope of future research.

\subsection{Limitations and Future Directions}
While the results of this study are promising, several limitations must be acknowledged. First, the study focused on a single neurosurgical dataset with a limited number of parameters. It is unclear whether reduced performance would be observed if this approach was applied to data with additional features. Second, the study evaluated the model’s ability to retain relationships between parameters through the correlation between KPS deterioration at discharge and the Landriel scale, and relationship between height, weight, and BMI. While these relationships are clinically relevant and preservation shows initial promise, neurosurgical datasets often include more bivariate and multivariate relationships. Further research is necessary to assess the preservation of additional relationships between parameters, including more nuanced interdependencies. Further validation of internal consistency, plausibility, and clinical relevance are similarly necessary. Third, while no generated dataset exhibited completely identical data to another, there was data overlap. It is unclear whether this is due to similarities in processing or cached data, despite the independence of trials and disabled memory settings. Finally, the performance on binary classification was still limited, and further work is necessary to improve data utility. Iterations of LLMs have previously shown improvements in performance on several tasks [29,30]; therefore, future versions of GPT-4o and other LLMs may also improve upon the presented results. \\
\\
The presented findings provide strong impetus for future research to continue developing LLM-based approaches to synthetic data generation. While this study involved a variety of pathologies to demonstrate generalizability, future work should also focus on particular neurosurgical subspecialties, diseases, and/or patient populations. Application toward datasets with additional parameters and more complex bivariate and multivariate statistical properties is warranted. Visualizing and comparing bivariate correlational heatmaps between synthetic and RWD should also be conducted in follow-up work. Further analyses may also include additional metrics to measure privacy preservation, including assessment of membership inference and attribute inference risks [31]. \\
\\
Alternative approaches to prompting may be explored through ablation studies and iterative prompting. Follow-up prompts may instruct the model to adjust properties after analyzing a generated dataset (i.e., fine-tuning), to evaluate its effect on fidelity and utility. Given that some generations exhibited greater fidelity than others, selecting the best performing generation over successive trials for the intended application—similar to research using other generative techniques—could establish a standard approach to this method [32]. Finally, future research should evaluate the feasibility of using synthetic data in place of providing RWD from clinical research for secondary analyses. Clinical data is often inaccessible for secondary analyses [21,33], and privacy-preserving, high-fidelity synthetic data may offer a viable and scalable alternative.

\section{Conclusions}
GPT-4o demonstrated a promising ability to generate high-fidelity synthetic neurosurgical data. These findings also indicate that data synthesized with GPT-4o can effectively augment clinical data with small sample sizes, and train ML models for prediction of neurosurgical outcomes. Notably, this zero-shot prompting approach requires no pre-training, fine-tuning, or access to RWD, offering a potential solution to critical limitations in data availability for neurosurgical research. Further investigation is necessary to improve the preservation of distributional characteristics and boost classifier performance. \\
\newpage

\section{References}
[1] Huang J, Shlobin NA, DeCuypere M, Lam SK. Deep learning for outcome prediction in neurosurgery: a systematic review of design, reporting, and reproducibility. Neurosurgery. 2022;90(1):16-38. doi:10.1227/NEU.0000000000001736\\

[2] Zlochower A, Chow DS, Chang P, Khatri D, Boockvar JA, Filippi CG. Deep learning AI applications in the imaging of glioma. Topics in Magnetic Resonance Imaging. 2020;29(2):115-00. doi:10.1097/RMR.0000000000000237\\

[3] Hashimoto DA, Rosman G, Rus D, Meireles OR. Artificial intelligence in surgery: promises and perils. Annals of Surgery. 2018;268(1):70-76. doi:10.1097/SLA.0000000000002693\\

[4] Pavlenko E, Strech D, Langhof H. Implementation of data access and use procedures in clinical data warehouses. A systematic review of literature and publicly available policies. BMC Med Inform Decis Mak. 2020;20(1):157. doi:10.1186/s12911-020-01177-z\\

[5] Newgard CD, Lewis RJ. Missing data: how to best account for what is not known. JAMA. 2015;314(9):940. doi:10.1001/jama.2015.10516\\

[6] Wartenberg D, Thompson WD. Privacy versus public health: the impact of current confidentiality rules. Am J Public Health. 2010;100(3):407-412. doi:10.2105/AJPH.2009.166249\\

[7] Sugiyama T, Sugimori H, Tang M, Fujimura M. Artificial intelligence for patient safety and surgical education in neurosurgery. JMA J. 2025;8(1):76-85. doi:10.31662/jmaj.2024-0141\\

[8] Tudur Smith C, Nevitt S, Appelbe D, et al. Resource implications of preparing individual participant data from a clinical trial to share with external researchers. Trials. 2017;18(1):319. doi:10.1186/s13063-017-2067-4\\

[9] Willemink MJ, Koszek WA, Hardell C, et al. Preparing medical imaging data for machine learning. Radiology. 2020;295(1):4-15. doi:10.1148/radiol.2020192224\\

[10] Legido-Quigley C, Wewer Albrechtsen NJ, Bæk Blond M, et al. Data sharing restrictions are hampering precision health in the European Union. Nat Med. Published online January 17, 2025. doi:10.1038/s41591-024-03437-1\\

[11] Sheehan JP, Michalopoulos GD, Katsos K, Bydon M, Asher AL. The NeuroPoint Alliance SRS \& Tumor QOD registries. J Neurooncol. 2024;166(2):257-264. doi:10.1007/s11060-023-04553-7

[12] Asher AL, Khalafallah AM, Mukherjee D, et al. Launching the quality outcomes database tumor registry: rationale, development, and pilot data. Journal of Neurosurgery. 2022;136(2):369-378. doi:10.3171/2021.1.JNS201115\\

[13] McGirt MJ, Speroff T, Dittus RS, Harrell FE, Asher AL. The National Neurosurgery Quality and Outcomes Database (N2QOD): general overview and pilot-year project description. Neurosurg Focus. 2013;34(1):E6. doi:10.3171/2012.10.FOCUS12297\\

[14] Asher AL, Haid RW, Stroink AR, et al. Research using the Quality Outcomes Database: accomplishments and future steps toward higher-quality real-world evidence. Journal of Neurosurgery. 2023;139(6):1757-1775. doi:10.3171/2023.3.JNS222601\\

[15] Asher AL, Speroff T, Dittus RS, et al. The national neurosurgery quality and outcomes database (N2QOD): a collaborative north american outcomes registry to advance value-based spine care. Spine. 2014;39:S106-S116. doi:10.1097/BRS.0000000000000579\\

[16] Bellovin SM, Dutta PK, Reitinger N. Privacy and synthetic datasets. SSRN Journal. Published online 2018. doi:10.2139/ssrn.3255766\\

[17] Jordon J, Szpruch L, Houssiau F, et al. Synthetic data -- what, why and how? Published online 2022. arXiv:2205.03257 [cs.LG]\\

[18] Goodfellow IJ, Pouget-Abadie J, Mirza M, et al. Generative adversarial networks. Published online 2014. arXiv:1406.2661 [stat.ML]\\

[19] Kingma DP, Welling M. Auto-encoding variational bayes. Published online 2013. arXiv:1312.6114 [stat.ML]\\

[20] Rajotte JF, Bergen R, Buckeridge DL, El Emam K, Ng R, Strome E. Synthetic data as an enabler for machine learning applications in medicine. iScience. 2022;25(11):105331. doi:10.1016/j.isci.2022.105331\\

[21] Azizi Z, Zheng C, Mosquera L, Pilote L, El Emam K. Can synthetic data be a proxy for real clinical trial data? A validation study. BMJ Open. 2021;11(4):e043497. doi:10.1136/bmjopen-2020-043497\\

[22] Koetzier LR, Wu J, Mastrodicasa D, et al. Generating synthetic data for medical imaging. Radiology. 2024;312(3):e232471. doi:10.1148/radiol.232471\\

[23] Chen D, Yu N, Zhang Y, Fritz M. GAN-leaks: a taxonomy of membership inference attacks against generative models. In: Proceedings of the 2020 ACM SIGSAC Conference on Computer and Communications Security. ACM; 2020:343-362. doi:10.1145/3372297.3417238\\

[24] Barr AA, Quan J, Guo E, Sezgin E. Large language models generating synthetic clinical datasets: a feasibility and comparative analysis with real-world perioperative data. Front Artif Intell. 2025;8:1533508. doi:10.3389/frai.2025.1533508\\

[25] Xu L, Skoularidou M, Cuesta-Infante A, Veeramachaneni K. Modeling tabular data using conditional gan. Published online 2019. arXiv:1907.00503 [cs.LG]\\

[26] Ferroli P, Vetrano IG, Schiavolin S, et al. Dataset related to article “brain tumor resection in elderly patients potential factors of postoperative worsening in a predictive outcome model.” Published online January 19, 2022. doi:10.5281/ZENODO.5879094\\

[27] Ferroli P, Vetrano IG, Schiavolin S, et al. Brain tumor resection in elderly patients: potential factors of postoperative worsening in a predictive outcome model. Cancers. 2021;13(10):2320. doi:10.3390/cancers13102320\\

[28] Hayes J, Melis L, Danezis G, De Cristofaro E. LOGAN: membership inference attacks against generative models. Published online 2017. arXiv:1705.07663 [cs.CR]\\

[29] Meyer A, Riese J, Streichert T. Comparison of the performance of GPT-3.5 and GPT-4 with that of medical students on the written german medical licensing examination: observational study. JMIR Med Educ. 2024;10:e50965. doi:10.2196/50965\\

[30] Rosoł M, Gąsior JS, Łaba J, Korzeniewski K, Młyńczak M. Evaluation of the performance of GPT-3.5 and GPT-4 on the Polish medical final examination. Sci Rep. 2023;13(1):20512. doi:10.1038/s41598-023-46995-z\\

[31] Yan C, Yan Y, Wan Z, et al. A Multifaceted benchmarking of synthetic electronic health record generation models. Nat Commun. 2022;13(1):7609. doi:10.1038/s41467-022-35295-1\\

[32] Eunice HW, Hargreaves CA. Simulation of synthetic diabetes tabular data using generative adversarial networks. Clin Med J. 2021;7(2): 49-59.\\

[33] Bergeat D, Lombard N, Gasmi A, Le Floch B, Naudet F. Data sharing and reanalyses among randomized clinical trials published in surgical journals before and after adoption of a data availability and reproducibility policy. JAMA Netw Open. 2022;5(6):e2215209. doi:10.1001/jamanetworkopen.2022.15209\\

\end{document}